\pdfoutput=1

\documentclass[11pt]{article}

\usepackage{multirow}
\usepackage{booktabs}
\usepackage[table]{xcolor}
\usepackage{graphicx}

\usepackage{amsmath,amsfonts,bm}

\def\eqref#1{equation~\ref{#1}}

\def\1{\bm{1}}

\def\vh{{\bm{h}}}

\def\vw{{\bm{w}}}
\def\vx{{\bm{x}}}
\def\vy{{\bm{y}}}

\def\mS{{\bm{S}}}
\def\mT{{\bm{T}}}

\DeclareMathAlphabet{\mathsfit}{\encodingdefault}{\sfdefault}{m}{sl}
\SetMathAlphabet{\mathsfit}{bold}{\encodingdefault}{\sfdefault}{bx}{n}

\DeclareMathOperator*{\argmax}{arg\,max}

\usepackage[]{ACL2023}

\usepackage{enumitem}

\usepackage{times}
\usepackage{latexsym}

\usepackage{amsmath}
\usepackage{amsthm}

\usepackage[T1]{fontenc}

\usepackage[utf8]{inputenc}

\usepackage{microtype}

\usepackage{inconsolata}

\usepackage{amsmath}

\usepackage[linesnumbered,ruled,vlined]{algorithm2e}
\usepackage{algpseudocode}
\newtheorem{theorem}{Theorem}[section]
\newtheorem{proposition}[theorem]{Proposition}

\title{Filtered Semi-Markov CRF}

\author{Urchade Zaratiana$^{12}$, Nadi Tomeh$^2$, Niama El Khbir$^{2}$, \\ \textbf{Pierre Holat}$^{12}$, \textbf{Thierry Charnois}$^{2}$ \\
$^1$ FI Group, 
$^2$ LIPN, CNRS UMR 7030, France \\{\tt zaratiana@lipn.fr} \\ \texttt{https://github.com/urchade/Filtered-Semi-Markov-CRF}}

\begin{document}
\maketitle

\begin{abstract} Semi-Markov CRF has been proposed as an alternative to the traditional Linear Chain CRF for text segmentation tasks such as Named Entity Recognition (NER). Unlike CRF, which treats text segmentation as token-level prediction, Semi-CRF considers segments as the basic unit, making it more expressive. However, Semi-CRF suffers from two major drawbacks: (1) quadratic complexity over sequence length, as it operates on every span of the input sequence, and (2) inferior performance compared to CRF for sequence labeling tasks like NER. In this paper, we introduce Filtered Semi-Markov CRF, a variant of Semi-CRF that addresses these issues by incorporating a filtering step to eliminate irrelevant segments, reducing complexity and search space. Our approach is evaluated on several NER benchmarks, where it outperforms both CRF and Semi-CRF while being significantly faster. The implementation of our method is available on \href{https://github.com/urchade/Filtered-Semi-Markov-CRF}{Github}.
\end{abstract}

\section{Introduction}

Sequence segmentation, the process of dividing a sequence into distinct, non-overlapping segments, has various applications, including Named Entity Recognition and Chinese Word Segmentation \cite{tjong-kim-sang-de-meulder-2003-introduction, li-yuan-1998-chinese}. In the past, this task has been approached as a sequence labeling problem using pre-defined templates, such as the BIO and BILOU schemes \cite{ratinov-roth-2009-design}. The Conditional Random Field (CRF) \cite{crf} has become a popular method for sequence labeling problems due to its ability to model the dependency between adjacent token tags. However, the CRF model may not efficiently capture the underlying structure of the sequence, as it is limited to modeling relationships between individual tokens rather than segments.

The Semi-Markov CRF \cite{semicrf} has been proposed as a variant of the CRF, allowing for the incorporation of higher-level segment features, such as segment width. While the Semi-CRF allows for a more natural approach to sequence segmentation, it suffers from slower learning and inference due to its quadratic complexity with respect to the sequence length. Additionally, the Semi-CRF often underperforms CRF, showing only marginal improvements in some cases \cite{Liang2005SemiSupervisedLF, Daum2005LearningAS,andrew-2006-hybrid}, which can be attributed to the Semi-CRF's significantly larger solution space, complicating the search for optimal solutions.

To address the limitations of Semi-CRF, we propose Filtered Semi-CRF, which introduces a filtering step to prune irrelevant segments using a lightweight local segment classifier. By leveraging transformer-based features, such as BERT \citep{devlin-etal-2019-bert}, this classifier can identify high-quality candidate segments. Consequently, the task of the Semi-CRF is simplified to selecting the best segments from the pool of high-quality candidates. Our experiments demonstrate that this filtering step not only accelerates the decoding process but also improves the overall model performance.

Although pruning techniques have been applied to accelerate parsing algorithms \citep{roark-hollingshead-2008-classifying, bodenstab-etal-2011-beam}, they often involve a trade-off between accuracy and inference speed. In contrast, our filtering approach is learned jointly and collaboratively with the Semi-CRF during training, resulting in a model that not only increases efficiency but also improves overall performance.

When evaluated on Named Entity Recognition, our model significantly outperforms both the CRF and Semi-CRF, achieving F1 score improvements of up to 2.5 and 1.1 points, respectively, on the CoNLL 2003 dataset. Additionally, our model also accelerates the decoding process to a speed that can be up to 20 times and 137 times faster than CRF and Semi-CRF, respectively.

\section{Background}

\subsection{Probabilistic structured predictor} In this paper, we aim to produce a structured output $\vy$ given an input sequence $\vx$. To assess the compatibility between the input and output, we employ a parameterized score function $\mS_{\theta}(\vy|\vx)$. The probability of a structure $\vy$ given $\vx$ is computed as follows:
\begin{align}
    p_{\theta}(\vy|\vx) = \frac{\exp \mS_{\theta}(\vy|\vx)}{\sum_{\vy' \in \mathcal{Y}(\vx)} \exp \mS_{\theta}(\vy'|\vx)}
    \label{eq:seg_pr}
\end{align}

where $\mathcal{Y}(\vx)$ represents the set of all possible outputs for $\vx$, and the denominator serves as a normalization constant, referred to as the partition function, denoted by $\mathcal{Z}_{\theta}(\vx)$.

\paragraph{Training} During training, the goal is to update the model's parameters $\theta$ to maximize the likelihood of the training data. The loss function for a pair of data points $(\vx, \vy)$ is computed as follows:
\begin{align}
\begin{split}
    \mathcal{L}(\vx, \vy) &= - \log p_{\theta}(\vy|\vx) \\
    &= - \mS_{\theta}(\vy|\vx) + \log \mathcal{Z}_{\theta}(\vx)
\end{split}
\end{align}
This loss function can be optimized using a stochastic gradient descent algorithm on the training data. Computing the partition function $\mathcal{Z}_{\theta}(\vx)$ can be challenging when the output space is large, but it can be calculated efficiently using dynamic programming in some cases.

\paragraph{Inference} During inference, the goal is to produce the most likely output:
\begin{align}
    \vy^* = \argmax_{\vy \in \mathcal{Y}(\vx)} \mS_{\theta}(\vy|\vx)
\end{align}
All the models we present in this paper follow this type of probabilistic modeling. For the remainder of this paper, we omit the dependency on $\theta$ for better readability.

\subsection{Linear Chain CRF}

The Linear-Chain CRF \citep{crf} is a sequence labeling model that assigns a label to each token in the input sequence, taking into account dependencies between adjacent labels. The score function of the CRF has the following form:
\begin{align}
\mS(\vy|\vx) = \sum_{i=1}^{|\vx|}\boldsymbol{\psi}(y_i|\vx)+\sum_{i=2}^{|\vx|}\mT[y_{i-1},y_{i}]
\end{align}

Here, $\boldsymbol{\psi}(y_i|\vx) \in \mathbb{R}$ is the sequence label score at position $i$, and  $\mT \in \mathbb{R}^{|Y| \times |Y|}$ is a learnable label transition matrix. The partition function is computed using the Forward algorithm and the Viterbi algorithm \citep{viterbi_forward} is used to determine the optimal labeling, both with a computational complexity of $\mathcal{O}(L|Y|^2)$ (More details in Appendix \ref{A:2}).

\subsection{Semi-Markov CRF \label{sec:Semi-CRF}}
The Semi-CRF, proposed by \citep{semicrf}, operates at the segment level and allows for the modeling of features that cannot be captured by traditional linear-chain CRFs. It produces a segmentation $\vy$ of length $M$ for an input sequence $\vx$ of length $L$ ($L \geq M$). A segmentation $\vy = \{s_1, \ldots, s_M\} \in \mathcal{Y}(\vx)$ satisfies the following properties:
\begin{itemize}
\item Each segment $s_k = (i_k, j_k, l_k) \in \vy$ consists of a start position $i_k$, an end position $j_k$, and a label $l_k \in Y$.
\item The segments have positive lengths and completely cover the input sequence positions $1, \ldots, L$ \underline{\textit{without overlapping}}. In other words, the start and end positions satisfy $i_1=1$, $j_M=L$, and for every $j_k$ and $i_k$ we have $1 \leq i_k \leq j_k \leq L$, and $i_{k+1} = j_k + 1$.
\end{itemize}

\label{par:ex} Consider a sentence from a Named Entity Recognition (NER) task: "\textit{Alain Farley} works at \textit{McGill University}". It would be segmented as $\vy$=$\text{[(1,2,\texttt{PER}), (3,3,\texttt{O}), (4,4,\texttt{O}), (5,6,\texttt{ORG})]}$, considering assumption from \citet{semicrf} that non-entity segments (referred to as \texttt{O} or \texttt{null} segments) have unit length. Furthermore, the Semi-CRF score function is defined as follows:
\begin{equation}
\displaystyle
\mS(\vy|\vx) = \sum_{k=1}^M \boldsymbol{\phi}(s_k |\vx) + \mT[l_{k-1},l_k]
\label{eq:score_semi}
\end{equation}

Here, $\boldsymbol{\phi}(s_k |\vx) \in \mathbb{R}$ represents the score of the $k$-th segment of $\vy$, and $\mT[l_{k-1},l_k]$ denotes the label transition score. Additionally, $\mT[l_0,l_1] = 0$. The partition function of the Semi-CRF can be computed in polynomial time using a modified version of the Forward algorithm and the segmental Viterbi algorithm is used to compute optimal segmentation (Appendix \ref{A:3} for details). The computational complexity of the Semi-CRF increases quadratically with both the sequence length and the number of labels, \textit{i.e} $\mathcal{O}(L^2|Y|^2)$.

\begin{figure*}
\centering
\includegraphics[width=1\textwidth]{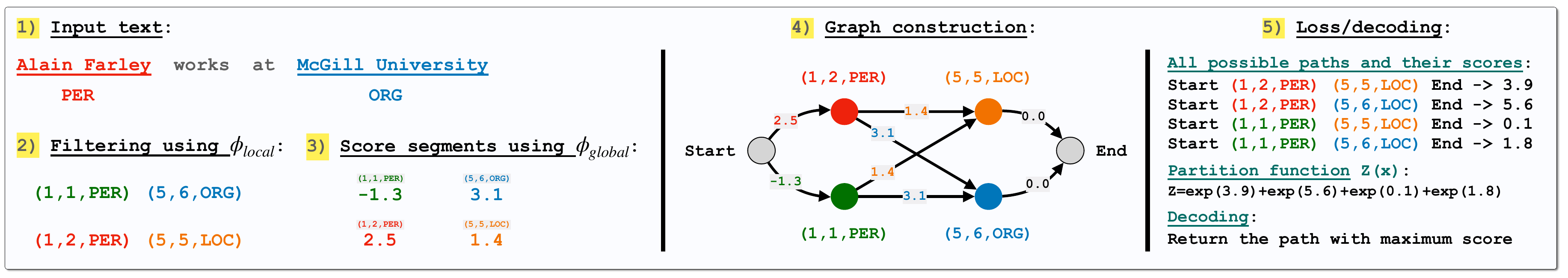}
    \caption{\textbf{Filtered Semi-CRF for NER}. The model takes as text sequence and output the best entity segments.}
    \label{fig:graph_con}
\end{figure*}

\subsection{Graph-based Formulation of Semi-CRF \label{sec:Semi-CRF_segment_path}}

In this section, we present a graph-based formulation of the Semi-CRF. As explained in \S~\ref{sec:Semi-CRF} , a sequence $\vx$ of length $L$ is divided into $M$ labeled segments $(i_k, j_k, l_k)$, with $i_k$, $j_k$ and $l_k$ denoting respectively the start position, end position and the label. We define a directed graph $\mathcal{G}(V_{\text{full}}, E)$, with $V_{\text{full}}$ its set of nodes composed of all possible segments $s_k$ of $\vx$:
\begin{align}
V_{\text{full}} = \bigcup_{i=1}^{L}\bigcup_{j=i}^{L}\bigcup_{l=1}^{\left| Y\right|} \{(i, j, l)\},
\end{align}
and an edge $s_{k'} \rightarrow s_k \in E$ exists if and only if the start position of $s_k$ immediately follows the end position of $s_{k'}$, i.e., $j_{k'}+1 = i_k$. The weight of an edge $s_{k'} \rightarrow s_k$ is defined as:
\begin{align}
w(s_{k'} \rightarrow s_k |\vx) = \phi(s_{k}|\vx)+\mT[l_{k'},l_k],
\end{align}

where $\phi(s_{k} |\vx)$ is the score of the segment $s_k$ and $\mT[l_{k'},l_k]$ is the label transition score. Moreover, Any directed path ${s_1, s_2,\ldots,s_M}$ of the graph $\mathcal{G}$ corresponds to a valid segmentation of $\vx$ if it verifies the segmentation properties described in \S~\ref{sec:Semi-CRF} . Additionally, the score of a valid path is computed as the sum of the edge scores, and is equivalent to the Semi-CRF score of the segmentation (\textit{Eq.} \ref{eq:score_semi}):
\begin{equation}
\resizebox{0.895\hsize}{!}{%
$\begin{aligned}
\mS(s_1,\ldots,s_M|\vx) &= \sum_{k=1}^M w(s_{k-1} \rightarrow s_k|\vx) \\
&= \sum_{k=1}^M \phi(s_k |\vx)+\mT[l_{k-1},l_k ]
\end{aligned}$
}
\end{equation}

The search for the best segmentation of the sequence $\vx$ is equivalent to finding the maximal weighted path of the graph $\mathcal{G}$ that starts at $i_1=1$ and ends at $j_M=L$. This search can be done using a generic shortest path algorithm such as \textit{Bellman-Ford}, whose complexity is of $L^3$. Nevertheless, taking into account the lattice structure of the problem, the Viterbi algorithm \citep{1054010, viterbi_forward} can achieve this while reducing the complexity to $L^2$.

\section{Filtered Semi-Markov CRF \label{sec:3}}

In this section,  we propose an alternative model to Semi-CRF, named Filtered Semi-CRF, which aims to address two fundamental weaknesses of the original model.
First, the Semi-CRF is not well-suited for long texts due to its \underline{\textit{quadratic complexity}} and the prohibitively large search space. Secondly, in tasks such as Named Entity Recognition (NER), where certain segments are labeled as \texttt{null} (representing non-entity segments), the Semi-CRF graph can create \underline{\textit{multiple redundant paths}}, all leading to the same set of entities. For instance, consider the scenario described in \S~\ref{sec:Semi-CRF}. In this scenario, multiple segmentations, such as $\vy$=$\text{[(1,2,\texttt{PER}), (3,3,\texttt{O}), (4,4,\texttt{O}), (5,6,\texttt{ORG})]}$ or $\vy$=$\text{[(1,2,\texttt{PER}), (3,4,\texttt{O}), (5,6,\texttt{ORG})]}$, would yield the same final set of labeled entities, specifically (1,2, \texttt{PER}) and  (5,6,\texttt{ORG}) in this case.
To remedy these shortcomings, our proposed model incorporates a filtering step that eliminates irrelevant segments using a lightweight local classifier. By leveraging transformer-based features, this classifier effectively selects high-quality candidate segments, significantly reducing the task of the Semi-CRF to merely choosing the best among already high-quality candidates.

\subsection{Filtering}
\paragraph{Local classifier \label{local_clf}} We first define the local classifier $\phi_{\text{local}}$ as a model that assigns a score to a labelled segment $s=(i, j, l)$ given an input sequence $\mathbf{x}$:
\begin{align}
    \phi_{\text{local}}(s=(i, j, l)|\vx) = {\vw_{l}}^{T} f(\vh_{i}, \ldots, \vh_{j})
    \label{eq:local_c}
\end{align}

where $\mathbf{h}_i \in \mathbb{R}^D$ is the token representation at position $i$ (computed by a pretrained transformer such as BERT), and $\mathbf{w}_l \in \mathbb{R}^D$ is a learnable weight associated with the label $l$. The function $f$ represents the segment featurizer, which aggregates token representations into a single feature representation. We found that a simple sum operation provides strong performance across settings.

\paragraph{Filtered graph \label{edge_gr}} The filtering consists in removing the segments $s_k = (i_k, j_k, l_k)$ for which $l_k = \argmax_l \phi_{local}(i_k, j_k, l | x)$ and $l_k = \texttt{null}$:
\begin{equation}
\resizebox{1\hsize}{!}{%
$
V = \left\{(i_k, j_k, l_k) \in V_{\text{full}} \ \middle|\ 
\begin{array}{l} l_k = \argmax_l \phi_{local}(i_k, j_k, l | x) \\ \land \; \\ l_k \neq \texttt{null} \end{array} \right\}$%
}
\label{eq:filtering}
\end{equation}
This new set of filtered nodes $V$ requires to define the set of edges $E$ differently from the definition of \S~\ref{sec:Semi-CRF_segment_path}. Thus, we propose to define the edges following \citet{mwis}: $\forall (s_{k'}, s_{k}) \in V^2$, $s_{k'} \rightarrow s_k \in E$ if $j_{k'} < i_{k}$ and there is no $s_{k^*} \in V$ such that $j_{k'} < i_{k^*}$ and  $i_{k^*} < j_{k}$. This definition means that $s_{k'} \rightarrow s_k$ is an edge if the start position of $s_k$ follows the end position of $s_{k'}$, and that no other segment lies completely in between these two positions $(j_{k'}, i_k)$. This formulation generalizes the Semi-CRF to graphs with missing segments. However, with missing segments, the starting and ending positions of segmentations do not necessarily verify $i_1=1$ and $j_M=L$. Thus, we simply add two terminal nodes $\texttt{start}$ and $\texttt{end}$, verifying: 
\begin{equation*}
\centering
\resizebox{1\hsize}{!}{%
$\begin{cases}
\texttt{start} \rightarrow s_k \in E \text{ \textit{iff} } \forall k' \neq \texttt{start}, s_{k'} \rightarrow s_k \notin E \\
s_k \rightarrow \texttt{end} \in E \text{ \textit{iff} } \forall k' \neq \texttt{end}, s_k \rightarrow s_{k'} \notin E
\end{cases}$%
}
\end{equation*}
In this context, a segmentation is simply a path in the graph starting at $\texttt{start}$ and ending at $\texttt{end}$ node (see Figure \ref{fig:graph_con}).  Referring back to the example in \S~\ref{sec:Semi-CRF}, the correct segmentation of "\textit{Alain Farley} works at \textit{McGill University}" using the Filtered Semi-CRF would be $\vy$=\text{[\texttt{start}, (1, 2, \texttt{PER}), (5, 6, \texttt{ORG}), \texttt{end}]}, where all remaining part of the segmentation are considered as having \texttt{null} labels.

\subsection{Segmentation scoring \label{seg_score}} In the filtered graph, the score of a segmentation, $\vy=\{\texttt{start}, s_1, \ldots, s_M, \texttt{end}\}$ is computed by summing its edge scores as for the Semi-CRF described in \S~\ref{sec:Semi-CRF_segment_path}:
\begin{equation}
\begin{split}
    \mS(\vy|\vx) &= \sum_{s_k \in \vy} w(s_{k'} \rightarrow s_k|\vx) \\
    &=\sum_{s_k \in \vy} \boldsymbol{\phi}_{global}(s_k|\vx)+\mT[l_{k'}, l_k]
\end{split}
\end{equation}

where $\phi_{global}$ is a model that computes score of the nodes/segments in the filtered graph, defined similarly as $\phi_{local}$ in \S~\ref{local_clf} and they share the same feature $f$. $\mT[l_{k'}, l_k]$ represents the transition score between the adjacent labels. By default, we set $w(\texttt{start} \rightarrow s_1)=\phi_{global}(s_1|\vx)$ and $w(s_{M} \rightarrow \texttt{end})=0$. See figure \ref{fig:graph_con} for a visual example.


\section{Training}
In this section, we present our FSemiCRF training which involves updating the whole model parameters to minimize the following loss function:
\begin{align}
\mathcal{L} = \mathcal{L}_{\textit{local}}+\mathcal{L}_{\textit{global}}
\end{align}

Here, $\mathcal{L}_{\textit{local}}$ and $\mathcal{L}_{\textit{global}}$ represent the filtering loss and the segmentation loss, respectively.

\subsection{Filtering loss} The filtering loss is the sum of the negative log-probability of all gold-labeled segments, $V^*$:
\begin{equation}
\resizebox{0.85\hsize}{!}{%
$\begin{split}
    \mathcal{L}_{\textit{local}} &= -  \sum_{\substack{(i,j,l^*)\in V^*}} \log p(i, j, l^*|\vx) \\
    & = -\sum_{\substack{(i,j,l)\in V^*}} \log \frac{\exp \phi_{local}(i, j, l|\vx)}{\sum_{l'} \exp \phi_{local}(i, j, l'|\vx)}
\end{split}$%
}
\label{eq:local}
\end{equation}

In practice, we assign a lower weight to the loss of null segments to account for the imbalanced nature of the task. For that, we down-weight the loss for the label $l=\texttt{null}$ by a ratio $\beta \in ]0, 1]$, tuned on the dev set.

\subsection{Segmentation loss} 
The segmentation loss is the negative log-likelihood of the gold path $\vy$ in the filtered graph:
\begin{align}
    \mathcal{L}_{\textit{global}} = - \mS(\vy|\vx) + \log \mathcal{Z}(\vx)
    \label{eq:lg}
\end{align}

$\mS(\vy|\vx)$ is the segmentation score as per \S~\ref{seg_score}, and the partition function $\mathcal{Z}(\vx)$, the sum of exponentiated scores for all valid paths in the graph from $\texttt{start}$ to $\texttt{end}$. It can be computed efficiently via a message-passing algorithm \citep{jordanw}:

\begin{algorithm}[!ht]
\DontPrintSemicolon
\SetNlSty{textbf}{(}{)}
\SetAlgoNlRelativeSize{-1}

{Topologically sort the nodes in $V$} 

$\alpha[\texttt{start}] = 1$ and $\alpha[k] = 0$ otherwise for $k\in V$

\ForAll{$k\neq \texttt{start}$ in $V$}{
    \ForAll{$k'$ such that $k'\rightarrow k \in E$}{
        $\alpha[k] \gets \alpha[k] + \alpha[k'] \exp\{w(s_{k'} \rightarrow s_k)|\vx\}$
    }
}

$\mathcal{Z}(\vx)=\alpha[\texttt{end}]$        
\caption{Computing $\mathcal{Z}(\vx)$} 
\label{alg:partition}
\end{algorithm}

In practice, this implementation of $\mathcal{Z}(\vx)$ can be unstable, thus, all computations were performed in log space to prevent issues of overflow or underflow. The complexity of the algorithm is $\mathcal{O}(|V|+|E|)$ as it performs a topological sort (which visits each node and edge once), and then iterates over each node and its incoming edges exactly once, performing constant time operations.


\paragraph{} \textit{During training}, we impose certain constraints to ensure that the gold segmentation $\vy$ forms a valid path in the \textit{filtered} graph (with nodes $V$), which is critical for maintaining a positive loss, i.e., $\log \mathcal{Z}(\mathbf{x}) > \mS(\mathbf{y}|\mathbf{x})$: 1) All segments in $V$ that do not overlap with at least one segment from the gold segmentation $\vy$ are excluded. 2) All segments from the gold segmentation, even those not initially selected in the filtering step, are included in $V$.

\subsection{Inference \label{p:inf}} During inference, the first step is to obtain the candidate segments $V$ through filtering, and then constructing the graph $\mathcal{G}(V, E)$ (see \S~\ref{edge_gr}). The final results is obtained by identifying the path, from \texttt{start} to \texttt{end}, in the graph that has the highest score. We achieve this by using a max-sum dynamic programming algorithm, which has a similar structure to Algorithm \ref{alg:partition}:

\begin{algorithm}[!ht]
\DontPrintSemicolon
\SetNlSty{textbf}{(}{)}
\SetAlgoNlRelativeSize{-1}

{Topologically sort the nodes in $V$} 

$\delta[\texttt{start}] = 0$

\ForAll{$k\neq \texttt{start}$ in $V$}{
    $\delta[k] = \max_{\substack{k' \\ (k'\rightarrow k) \in E}} \delta[k'] + w(s_{k'}\rightarrow s_k|\vx)$
}

$\vy^* = \texttt{Traced}(\delta[\texttt{end}])$

\caption{Decoding}
\label{alg:decoding}
\end{algorithm}

The highest scoring path $\vy^*$, represented by $\text{argmax}_{\vy}\mS(\vy|\vx)$, is identified by the path traced by $\delta[\texttt{end}]$, which can be obtained through backtracking. This algorithm has a computational complexity of $\mathcal{O}(|V|+|E|)$, the same as that of computing the partition function $\mathcal{Z}(\vx)$ in Algorithm \ref{alg:partition}.

\begin{figure*}
    \centering
    \includegraphics[width=2\columnwidth]{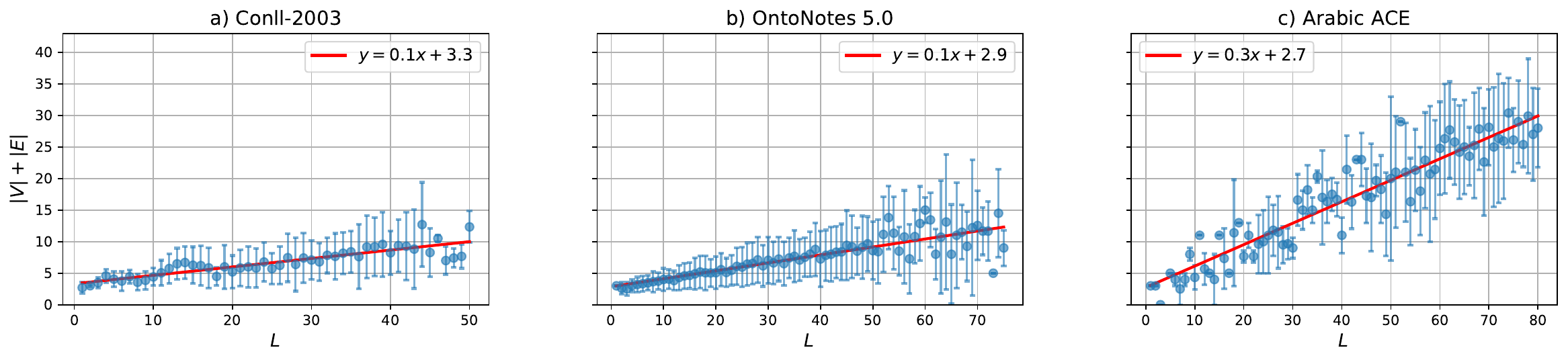}
    \caption{\textbf{Empirical complexity analysis}. We conducted an empirical complexity analysis using trained Filtered Semi-CRF models. The plot showcases the relationship between the size of the filtered graph ($|V|+|E|$) and the input sequence length $L$ on three NER datasets. As the length of the input sequence increases, the graph size seems to grow in a linear fashion.}
\label{fig:com}
\end{figure*}

\begin{figure*}
    \centering
\includegraphics[width=2\columnwidth]{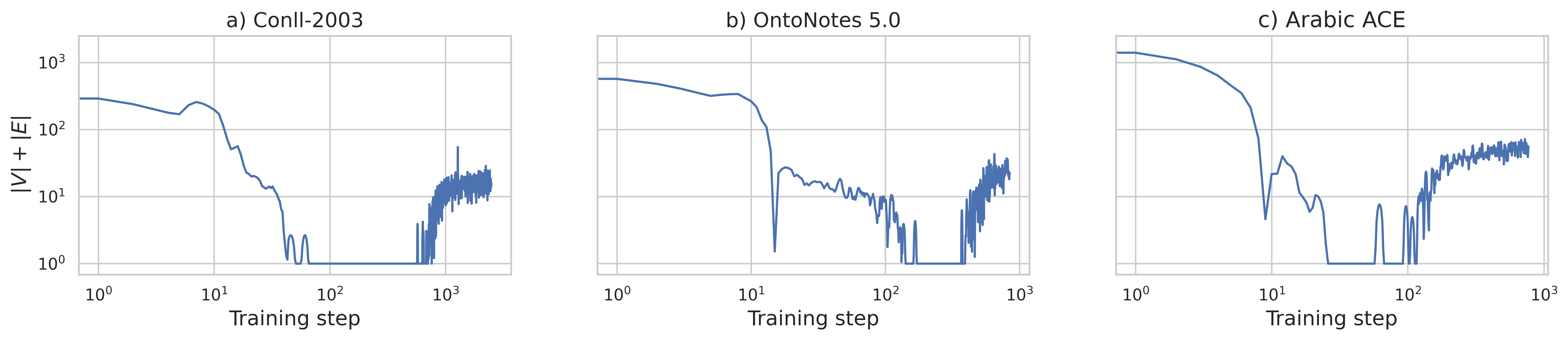}

\caption{\textbf{Graph Size during Training.} The graph size ($|V|+|E|+1$) undergoes three stages during training: 1) initially large when the filtering classifier is untrained, 2) decreasing in the second stage as most of segments in the training set have a \texttt{null} label (biasing the classifier toward this label), and 3) increasing again as the classifier improves, better aligning with the training dataset statistics.}
\label{fig:train}
\end{figure*}

\subsection{Complexity analysis \label{complexity:an}}
In this section, we analyze the complexity of the algorithms \ref{alg:partition} and \ref{alg:decoding}, $\mathcal{O}(|V|+|E|)$, as a function of the input sequence length $L$. Note that the size of $V$ does not depend on the number of labels $|Y|$ since there is at most one label per segment due to the filtering step in equation \ref{eq:filtering}. 

\begin{proposition}
The number of nodes in a Semi-CRF graph (as described in \S~\ref{sec:Semi-CRF_segment_path}) with an input length of $L$ is given by $\frac{L(L+1)}{2}$.
\end{proposition}

\begin{proposition}
The number of edges in a Semi-CRF graph (as described in \S~\ref{sec:Semi-CRF_segment_path}) with an input length of $L$ is given by $\frac{L(L-1)(L+1)}{6}$.
\end{proposition}

\noindent We employ these propositions to determine the complexity of the Filtered Semi-CRF model in the following. The proofs for these propositions can be found in Appendix \S~\ref{sec:proof}.

\paragraph{Worst case complexity}
In the worst case scenario, the filtering model $\boldsymbol{\phi}_{local}$ does not filter any segments, resulting in all segments being retained. By utilizing Propositions 3.1 and 3.2, we can deduce that in the worst case, $\mathcal{O}(|V|)=\mathcal{O}(L^2)$ and $\mathcal{O}(|E|)=\mathcal{O}(L^3)$. This implies that the complexity of our algorithm in the worst case is cubic with respect to the sequence length $L$, as $\mathcal{O}(|V|+|E|)=\mathcal{O}(L^3)$. However, it is worth noting that in this worst case scenario, the resulting graph is the Semi-CRF and the complexity can be reduced to $L^2$ by utilizing the Forward algorithm during training and the Viterbi algorithm during inference (\textit{Eq.} \ref{part:semicrf} and \textit{Eq.} \ref{vit:crf}).

\paragraph{Best Case Complexity}
In the ideal scenario, the filtering process is optimal, resulting in the number of nodes in the graph $|V|$ being equal to the true number of non-\texttt{null} segments in the input sequence, denoted by $\mathcal{S}$. Furthermore, since $\mathcal{S}$ does not contain overlapping segments, $|\mathcal{S}| \leq L$ with $|\mathcal{S}| = L$ if all segments in $\mathcal{S}$ have unit length and cover the entire sequence, i.e., $\mathcal{S}=\{(i, i, l_i)|i=1\ldots L, l_i \neq \texttt{null} \}$. Additionally, $|E| = |\mathcal{S}| - 1 \leq L - 1$ as optimal filtering implies that the path number is unique. As a result, in this best case scenario, the complexity of the algorithm is linear with respect to the sequence length $L$, i.e., $\mathcal{O}(L)$.

\paragraph{Empirical Analysis \label{emp_analis}}
In this study, we assess our model's empirical complexity by examining the correlation between the graph size ($|V|+|E|$) and the input sequence length $L$. We use three popular NER datasets for this analysis - CoNLL-2003, OntoNotes 5.0, and Arabic ACE. Our findings (shown in Figure \ref{fig:com}) indicate a linear increase in the graph size as the sequence length increases. Interestingly, the graph size always stays smaller than the sequence length. This suggests that in practice, the computational complexity of the FSemiCRF model is at worst, $\mathcal{O}(L)$. However, during the initial stages of model training, the graph size may be large because the filtering model, which is responsible for reducing the graph size, is not fully trained, as depicted in Figure \ref{fig:train}. But, the graph size decreases rapidly after a few training steps as the filtering classifier is improving.

\begin{table*}[!htbp]
\renewcommand{\arraystretch}{1.1}
\centering
\begin{tabular}{l|lll|lll|lll}
\toprule
Models & \multicolumn{3}{c}{CoNLL-2003} & \multicolumn{3}{c}{OntoNotes 5.0} & \multicolumn{3}{c}{Arabic ACE} \\
\cmidrule(r){2-4} \cmidrule(lr){5-7} \cmidrule(l){8-10}
& P & R & F & P & R & F & P & R & F \\
\midrule
\citet{Yu2020NamedER} & 93.7 & 93.3 & 93.5 & 91.1 & 91.5 & 91.3 & - & - & - \\
\citet{Yan2021AUG} & 92.61 & 93.87 & 93.24 & 89.99 & 90.77 & 90.38 & - & - & - \\
\citet{Zhu2022BoundarySF} & 93.61 & 93.68 & 93.65 & 91.75 & 91.74 & \textbf{91.74} & - & - & - \\
\citet{Shen2022ParallelIQ} & 93.29 & 92.46 & 92.87 & 91.43 & 90.73 & 90.96 & - & - & - \\
\citet{zaratiana-etal-2022-global} & 94.29 & 93.33 & 93.81 & 90.21 & 91.21 & 90.71 & 85.35    & 83.64    & \textbf{84.49}    \\
\citet{el-khbir-etal-2022-arabie} & - & - & - & - & - & - & 84.42 & 84.05 & 84.23 \\
\midrule
\multicolumn{10}{c}{Our experiments} \\
\midrule
\textbf{CRF} & 93.29 & 92.21 & 92.75 & 89.00 & 90.16 & 89.57 & 82.79 & 84.44 & 83.61 \\
\textbf{Semi-CRF} & 92.37 & 90.49 & 91.42 & 88.91 & 89.78 & 89.34 & 82.97 & 84.24 & 83.60 \\
\textbf{+ Unit size \texttt{null}$^\dagger$} & 92.08 & 91.41 & 91.74 & 89.17 & 89.76 & 89.47 & 83.35 & 83.62 & 83.48 \\
\textbf{FSemiCRF} & 94.72 & 93.09 & \textbf{93.89} & 90.69 & 91.31 & 91.00 & 83.43 & 85.51 & \textbf{84.46} \\
\textbf{-- w/o $\mathcal{L}_{\textit{global}}$ (\ref{eq:lg})$^\dagger$} & 94.24 & 92.70 & 93.46 & 90.85 & 89.57 & 90.21 & 83.73 & 83.56 & 83.64 \\
\bottomrule
\end{tabular}
\caption{\textbf{Main Results}. All English models employ \texttt{bert-large-cased} as token representations for English datasets, except \citep{Yan2021AUG} that uses \texttt{bart-large}. $^\dagger$ See the ablation study (\S~\ref{ablation}) for details.}
\label{tab:1}
\end{table*}

\section{Experimental setups}

\paragraph{Datasets and evaluation} We evaluate our models on on three diverse Named Entity Recognition (NER) datasets: CoNLL-2003 and OntoNotes 5.0, both English, and Arabic ACE (further details in Appendix \ref{A:data}). We adopt the standard NER evaluation methodology, calculating precision (P), recall (R), and F1-score (F), based on the exact match between predicted and actual entities.

\paragraph{Hyperparameters} To produce contextual token representations, we used \texttt{bert-large-cased} \citep{devlin-etal-2019-bert} for both CoNLL-2003 and OntoNotes 5.0 datasets, and \texttt{bert-base-arabertv2} \citep{Antoun2020AraBERTTM} for Arabic ACE. For simplicity, we do not use auxiliary embeddings (eg. \textit{character embeddings}). All models are trained with \textit{Adam} optimizer \citep{kingma2017adam}. We employed a learning rate of \texttt{2e-5} for the pre-trained parameters and a learning rate of \texttt{5e-4} for the other parameters. We used a batch size of 8 and trained for a maximal epoch of 15. We keep the best model on the validation set for testing. In this work, for all segment-based model, we restrict the segment to a maximum width $K$ to reduce complexity without harming the recall score on the training set (however some segments may be missed for the test set). By bounding the maximum width of the segments, we reduce the number of segments from $L^2$ to $LK$. Under this setup, the complexity of the Semi-Markov CRF becomes $O(LK|Y|^2)$. We implemented our model with PyTorch \citep{pytorch}. The pre-trained transformer models were loaded from HuggingFace's Transformers library, we used AllenNLP \citep{gardner-etal-2018-allennlp} for data preprocessing and the seqeval library \citep{seqeval} for evaluating the sequence labeling models. Our Semi-CRF implementation is based on pytorch-struct \citep{alex2020torchstruct}. We trained all the models on a server with V100 GPUs. 

\paragraph{Baselines}
We compare our Filtered Semi-CRF model against the CRF \citep{crf} and Semi-CRF \citep{semicrf}. Additionally, we include results from previous studies: BiaffineNER \citep{Yu2020NamedER}, BartNER \citep{Yan2021AUG}, Boundary Smoothing \citep{Zhu2022BoundarySF}, PIQN \citep{Shen2022ParallelIQ}, GSS \citep{zaratiana-etal-2022-global} and ArabIE \cite{el-khbir-etal-2022-arabie}. For English datasets, models use \texttt{bert-large-case} (except BartNER with \texttt{bart-large}). Our model for Arabic data utilizes \texttt{bert-base-arabertv2}.

\section{Main results}
\paragraph{FSemiCRF vs. CRF and Semi-CRF} As shown in Table \ref{tab:1}, our FSemiCRF model outperforms both the CRF and Semi-CRF reference models in all datasets, validating its effectiveness.  The Semi-CRF model, while providing competitive results, often lags behind, either matching or slightly underperforming the CRF model. This observation is in line with the findings of \citet{Liang2005SemiSupervisedLF}.


\paragraph{Comparison to pior works} In our work, we mainly compare our approach with previous work that we consider comparable, i.e. that uses sentence-level context and the same backbone model. As shown in the Table \ref{tab:1}, on all datasets, we found that our FSemiCRF achieves competitive results on all the datasets. For example, our approach outperforms a span-based model we proposed earlier \cite{zaratiana-etal-2022-global}, which uses the Maximum weighted independent set to select the best spans.

\begin{table*}[t]
\renewcommand{\arraystretch}{1.2}
\centering
\resizebox{\textwidth}{!}{%
\begin{tabular}{@{}l|ccc|ccc|ccc@{}}
\toprule
 & \multicolumn{3}{c|}{\textit{CoNLL-2003 (|Y| = 4)}} & \multicolumn{3}{c|}{\textit{OntoNotes 5.0 (|Y| = 18)}} & \multicolumn{3}{c}{\textit{Arabic ACE (|Y| = 7)}} \\ 
\cmidrule(lr){2-4} \cmidrule(lr){5-7} \cmidrule(lr){8-10}
& CRF & Semi-CRF & FSemiCRF & CRF & Semi-CRF & FSemiCRF & CRF & Semi-CRF & FSemiCRF \\ 
\midrule
Scoring & 3.9 &  3.9 &  3.9 & 4.8 &  4.9 &  4.9 & 8.1 &  8.3 &  8.3 \\
Decoding & 2.7 & 3.7 &  0.2 & 4.4 & 27.5 &  0.2 & 6.0 & 10.1 &  0.3 \\ \rowcolor{blue!10}
Decoding Speedup & 1.3x & 1.0x &  \textbf{18.5x} & 6.2x & 1.0x &  \textbf{137x} & 1.7x & 1.0x &  \textbf{33.7x} \\
\midrule
Overall & 6.6 & 7.6 &  4.1 & 9.2 & 32.4 &  5.1 & 14.1 & 18.4 &  8.6 \\ \rowcolor{blue!10}
Overall Speedup & 1.1x & 1.0x &  \textbf{1.8x} & 3.5x & 1.0x &  \textbf{6.3x} & 1.30x & 1.0x &  \textbf{2.1x} \\
\bottomrule
\end{tabular}%
}
\caption{\textbf{Inference Wall Clock Time} (lower is better). Comparison of required wall-clock time for the scoring (tokens for CRF, segments for Semi-CRF/FSemiCRF) and decoding processes, measured in \textit{milliseconds / sample}.}
\label{tab:3}
\end{table*}


\section{Ablation study \label{ablation}} 
\paragraph{Semi-CRF + Unit \texttt{null}} We study a variation of the Semi-CRF that only allows for the use of \texttt{null} labels for unit length segments. To do this, we simply modify the original Semi-CRF by eliminating/masking segmentation paths that contain null segments with a size greater than one. The motivation for this study is to fix the \textit{multiple redundant paths} problem of the Semi-CRF (\S~\ref{sec:3}). The results show that this approach improves performance on most of the datasets, but still does not perform as well as the other methods, thus validating the importance of segment filtering.

\paragraph{FSemiCRF w/o global loss}

We investigate the impact of removing the global loss (\textit{Eq.} \ref{eq:lg}) component on our FSemiCRF model, resulting in a local span classfication model. The results are presented in Table \ref{tab:1} (w/o $\mathcal{L}_{\textit{global}}$), and show that even without the global loss component, the model still performs competitively. However, including global loss consistently improves the overall scores of FSemiCRF across all the datasets.

\subsection{Efficiency analysis} 
This section focuses on the computational efficiency of different models, for both training and inference. For this experiment, all the models use a base size for the encoder to ensure a fair comparison.

\paragraph{Inference wall clock time}
The wall clock time analysis for scoring and decoding operations, summarized in Table \ref{tab:3}, highlights subtle differences in scoring times across all models. However, when it comes to decoding, FSemiCRF significantly outperforms both CRF and Semi-CRF models on all datasets. Notably, FSemiCRF achieves a remarkable 137x speedup over Semi-CRF on the OntoNotes 5.0. Overall, FSemiCRF demonstrates superior performance, being up to 6x and 2x faster than CRF and Semi-CRF, respectively.

\begin{figure}
    \centering
    \includegraphics[width=1\columnwidth]{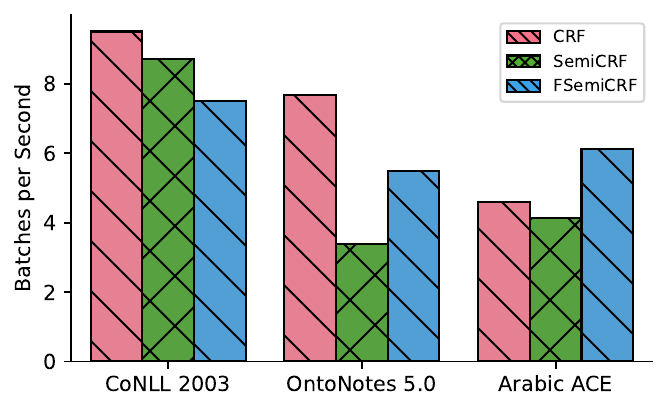}
    \vspace{-1.5em}
    \caption{\textbf{Training throughput} in \textit{batches per second}.}
    \label{fig:throu}
    \vspace{-1em}
\end{figure}

\paragraph{Training throughput} Figure \ref{fig:throu} presents the training throughput of the models, which measures the number of batches processed per second using a batch size of 8. It reveals that, in general, CRF is the fastest during training, with FSemiCRF following closely as the second fastest model. This can be attributed to the larger graph size of FSemiCRF during training, particularly in the early stages, which can potentially slow down the process, as discussed in the complexity analysis (\ref{emp_analis}). However, the differences in training performance between the models are not as pronounced as during inference.

\section{Related Work}

\paragraph{Linear-chain CRF} Numerous frameworks exist for text segmentation. The commonly used Linear-Chain CRF \citep{crf} treats this task as token-level prediction, training through sequence-level objectives and using the Viterbi algorithm \citep{1054010,Forney2010ViterbiA} for decoding. Variants have evolved from using handcrafted features \citep{crf,NIPS2006_24b43fb0,Roth2005IntegerLP} to automated feature learning through neural networks \citep{Do2010NeuralCR,Maaten2011HiddenUnitCR,kim-etal-2015-pre,huang2015bidirectional,Lample2016NeuralAF}. Higher order dependencies (Markov order $N>1$) have been explored for enhanced performance, but their adoption is limited due to complexity and marginal gains \citep{NIPS2009_94f6d7e0,JMLR:v15:cuong14a}.

\paragraph{Semi-Markov CRF} Semi-CRF \citep{semicrf} is an alternative operating at segment level, applied to tasks like Chinese word segmentation \citep{Kong2016SegmentalRN} and Named Entity Recognition \citep{semicrf, andrew-2006-hybrid, zhuo-etal-2016-segment, Liu2016ExploringSR, ye-ling-2018-hybrid}. It has the advantage of incorporating segment-level features but suffers from quadratic complexity and generally equivalent or marginally better performance than CRFs \citep{Liang2005SemiSupervisedLF,Daum2005LearningAS,andrew-2006-hybrid}. 


\paragraph{Dynamic Programming Pruning} Prior research has investigated the use of pruning techniques in dynamic programming to improve the efficiency of structured prediction tasks \citep{roark-hollingshead-2008-classifying,37844,bodenstab-etal-2011-beam,Vieira2017LearningTP}. These approaches aim to optimize runtime by selectively discarding hypotheses during inference. However, these methods often involve a trade-off between efficiency and performance. In contrast, our Filtered Semi-CRF model introduces a learned filtering step that collaboratively improves both efficiency and overall model performance.

\paragraph{Named Entity Recognition}
NER is an important task in Natural Language Processing and is used in many downstream information extraction applications such as relation extraction \citep{zaratiana2023an} and taxonomy construction \citep{10.1145/3219819.3220064,10.1007/978-3-031-01333-1_3}. Usually, NER tasks are designed as sequence labelling \citep{huang2015bidirectional,Lample2016NeuralAF,akbik-etal-2018-contextual} where the goal is to predict tagged sequence (eg. BIO tags). Recently, different approaches have been proposed to perform NER tasks that go beyond traditional sequence labelling. One approach that has been widely adopted is the span-based approach \citep{10.5555/3060832.3061024, fu-etal-2021-spanner, li2021empirical, zaratiana-etal-2022-global,zaratiana-etal-2022-gnner,zaratiana-etal-2022-named, lou-etal-2022-nested, Corro2022ADP} where the prediction is done in the span level instead of entity level. Futhermore, the use of the sequence-to sequence models for Named Entity Recognition has become popular recently. For instance, \citet{Yan2021AUG} uses the BART \citep{lewis2019bart} model to generate named entity using encoder-decoder with copy mechanism.

\section{Conclusion}
In this paper, we introduce Filtered Semi-CRF, a novel algorithm for text segmentation tasks. By applying our method to NER, we show substantial performance gains over traditional CRF and Semi-CRF models on several datasets. Additionally, our algorithm exhibits improved efficiency, speed, and scalability compared to the baselines. As future work, we plan to investigate the extension of Filtered Semi-CRF to nested segment structures.

\section*{Limitations}
While our Filtered Semi-CRF model offers several advantages, it also has limitations that should be considered:

\paragraph{Sensitivity to Filtering Quality}
The overall performance and efficiency heavily rely on the accuracy of the filtering process in identifying high-quality candidate segments. Inaccurate filtering or the introduction of errors during this step can adversely affect the model's performance.

\paragraph{Restriction to Non-overlapping Entities}
Our model is designed specifically for non-overlapping entity segmentation. It assumes that entities within the text do not overlap with each other. While this assumption is valid for many applications and datasets, scenarios exist where specific cases of entity overlap occurs, such as nested entities.

\section*{Acknowledgments} We are grateful to Joseph Le Roux for his valuable feedback on earlier versions of this paper. His insights and suggestions significantly enhanced the quality of this work. This work was granted access to the HPC/AI resources of [CINES/IDRIS/TGCC] under the allocation 2022AD011013096R1 made by GENCI.

\bibliography{custom}
\bibliographystyle{acl_natbib}

\appendix
\onecolumn

\section{Appendix}

\subsection{Proofs \label{sec:proof}}

\paragraph{Proposition 3.1}
\textit{The number of nodes in a Semi-CRF graph (as described in \S~\ref{sec:Semi-CRF_segment_path}) with an input length of $L$ is given by $\frac{L(L+1)}{2}$.}
\begin{proof}
Nodes are the enumeration of all segments (regardless of labels). Thus,
\begin{align}
V = \bigcup_{i=1}^L\bigcup_{j=i}^L (i, j) \Longrightarrow
    |V| &= \sum_{i=1}^L \sum_{j=i}^L 1 = \sum_{i=1}^L (L+1-i) \nonumber\\ 
    &= \sum_{i=1}^L (L+1) - \sum_{i=1}^L i = L(L+1) - \frac{L(L+1)}{2} \\ 
   |V| &=\frac{L(L+1)}{2} \nonumber
\end{align}
\end{proof}

\paragraph{Proposition 3.2}
\textit{The number of edges in a Semi-CRF graph (as described in \S~\ref{sec:Semi-CRF_segment_path}) with an input length of $L$ is given by $\frac{L(L-1)(L+1)}{6}$.}
\begin{proof}
We know that in the complete segment graph
\begin{enumerate}
    \item By definition, $(i_k, j_k) \rightarrow (i_{k'}, j_{k'}) \in E$ iff $j_k+1=i_{k'}$
    \item There are {\color{blue} $j_k$} segments ending at $j_k$ \textit{i.e} $|\bigcup_{i=1}^{j_k} (i, j_k)|={\color{blue} j_k}$
    \item There are {\color{red} $L-j_k$} segments starting at $i_{k'}$ \textit{i.e} $|\bigcup_{i=i_{k'}}^{L} (i_{k'}, i)|=L-i_{k'}+1={\color{red} L-j_k}$
\end{enumerate}

From 1, 2 and 3, we can deduce that there is ${\color{blue} j_k}({\color{red} L-j_k})$ segments starting at $i_{k'}$ and ending at $j_k$. Finally, the total number of edges of the graph is the sum over all $j_k$ from $0$ to $L$:
\begin{align}
    |E| &= \sum_{j_k=1}^L {\color{blue} j_k}({\color{red} L-j_k}) = L \sum_{j_k=1}^L j_k - \sum_{j_k=1}^L j_k^2 \nonumber\\ 
    &= L \frac{L(L+1)}{2} - \frac{L(L+1)(2L+1)}{6}
    = L(L+1)(\frac{L}{2} - \frac{2L+1}{6}) \\ 
   |E| &= \frac{L(L+1)(L-1)}{6} \nonumber
\end{align}
\end{proof}
\subsection{CRF \label{A:2}}
\paragraph{Partition function} The partition function $\mathcal{Z}(\vx)$ of the CRF \citep{crf} is computed using the forward algorithm, with $\alpha(1, y)=\boldsymbol{\psi}(y|\vx)$ and for $i=2\ldots L$:
\begin{align}
\label{part:crf}
\begin{split}
    \alpha(i, y)=\sum_{y' \in Y} \alpha(i-1, y') \exp\{\boldsymbol{\psi}(y|\vx) +  \mT[y',y]\} \\
    \mathcal{Z}(\vx) = \sum_{y \in Y} \alpha(L, y)
\end{split}
\end{align}

\paragraph{Decoding} The decoding of CRF is done with the Viterbi algorithm, with $\delta(1, y)=\boldsymbol{\psi}(y|\vx)$
\begin{align}
\label{vit:crf}
    \delta(i, y) = \max_{y' \in Y} \delta(i-1, y') + \boldsymbol{\psi}(y|\vx) +  \mT[y',y]
\end{align}
The best labeling is given by the path traced by $\max_{y \in Y}\delta(L, y)$. Both the computation of the partition function and the decoding of the CRF have a complexity of $\mathcal{O}(L|Y|^2)$.

\subsection{Semi-CRF \label{A:3}}
\paragraph{Partition function} The partition function of the Semi-CRF \citep{semicrf} $\mathcal{Z}(\vx)$ is computed using the following dynamic program (a modification of the forward algorithm) with $\alpha(0, :) = 1$ and $\alpha(m, :) = 0$ if $m<0$ and otherwise:
\begin{align}
\label{part:semicrf}
\begin{split}
      \alpha(m, y) = \sum_{d=1}^{L} \sum_{y' \in Y} \alpha(m-d, y') \exp\left\{\boldsymbol{\phi}((i=m-d+1, j=m, l=y) |\vx)+\mT[y',y]\right\} \\
    \mathcal{Z}(\vx) = \sum_{y \in Y} \alpha(L, y)  
\end{split}
\end{align}

\paragraph{Decoding}The decoding of the Semi-CRF is done with the segmental/Semi-Markov Viterbi algorithm with $\delta(0,:) = 0$ and $\delta(m,:) = -\infty$ if $m < 0$ and otherwise:
\begin{align}
\label{vit:semicrf}
    \delta(m,y)=\max_{\substack{y' \in Y \\  d=1\ldots L}} \delta(i-d,y') + \boldsymbol{\phi}((i=m-d+1, j=m, l=y) |\vx)+\mT[y',y]
\end{align}
The highest scoring segmentation is the path traced by $\max_{y \in Y}\delta(L, y)$. Both the computation of the partition function and the decoding of the Semi-CRF have a complexity of $\mathcal{O}(L^2|Y|^2)$.

\subsection{Datasets \label{A:data}} We evaluate our models on three diverse datasets of Named Entity Recognition. CoNLL-2003 \citep{tjong-kim-sang-de-meulder-2003-introduction} is a dataset from the news domain designed for extracting entities such as Person, Location, and Organization. OntoNotes 5.0 \citep{AB2/MKJJ2R_2013} is a large corpus comprising various kinds of text, including newswire, broadcast news, and telephone conversation, with a total of 18 different entity types, such as Person, Organization, Location, Product, or Date. Arabic ACE is the Arabic portion of the multilingual information extraction corpus, ACE 2005 \citep{ace05}. It includes texts from a wide range of genres, such as newswire, broadcast news, and weblogs, with a total of 7 entity types.

\end{document}